# Statistical Machine Translation for Indic Languages

Sudhansu Bala Das, Divyajoti Panda, Tapas Kumar Mishra, and Bidyut Kr. Patra

National Institute of Technology(NIT), Rourkela, Odisha, India

Indian Institute of Technology (IIT), Varanasi, Uttar Pradesh, India

**Abstract**

Machine Translation (MT) system generally aims at automatic representation of source language into target language retaining the originality of context using various Natural Language Processing (NLP) techniques. Among various NLP methods, Statistical Machine Translation (SMT) is a very popular and successful architecture used for both low as well as high-resource languages. SMT uses probabilistic and statistical techniques to analyze information and conversion. This paper canvasses about the development of bilingual SMT models for translating English to fifteen low-resource Indian Languages (ILs) and vice versa. At the outset, all 15 languages are briefed with a short description related to our experimental need. Further, a detailed analysis of Samanantar and OPUS dataset for model building, along with standard benchmark dataset (Flores-200) for fine-tuning and testing, is done as a part of our experiment. Different preprocessing approaches are proposed in this paper to handle the noise of the dataset. To create the system, MOSES open-source SMT toolkit is explored. "Distance" reordering is utilized with the aim to understand the rules of grammar and context-dependent adjustments through a phrase reordering categorization framework. In our experiment, the quality of the translation is evaluated using standard metrics such as BLEU, METEOR, and RIBES.

# 1 Introduction

Technology reaches new heights through its journey from the origins of ideas to their full-scale practical implementation. One such journey is heading towards elimination of language barrier in order to establish a seamless social communication in every domain. In this regard, advancement on relevant fields such as Natural Language Processing (NLP), Machine Learning (ML) and Artificial Intelligence (AI) based Language Modelling (LM) significantly contributes for evolving a flawless automatic Machine Translation (MT) system (Dorr et al ( 2004)). Irrespective of various heuristic approaches to maintain both lexical and contextual interpretation of source language(s) onto the translated target language(s), it is still challenging to cope with required fluency, adequacy, accent, and overall accuracy (Chapelle et al ( 2010)). However, it is feasible with the advent of modern NLP (AI-based) approaches wherein a high-quality and high resource (i.e. large quantity of corpora available) parallel corpus (translation pairs in source and target languages) is required to train a good translation system. Hence, for high-resource languages having massive digital footprint across the globe, MT systems prove to be quite efficient with adequate training. On the other hand, it becomes very complicated for low-resource languages suffering from universal recognition and scanty digital presence. Such imbalance often leads to poor-quality translation in presence of low-resource language(s) in the form of either target or source. Therefore, MT systems need to understand the syntax (rules to combine words), semantics (meaning of words and combinations), and morphology (rules to cover morphemes - smallest meaningful units - into words) of such low-resource languages (Somers (2011)).
Based on the heuristic paradigms, MT models are classified into rule-based (RBMT), example-based (EBMT), statistical (SMT), and neural (NMT) systems (Tripathi et al (2010)). Each has its own advantages and disadvantages. RBMT models follow a set of rules to define a language and the interaction between different linguistic devices (words, phrases, sentences) in the language (Jussà et al (2012), Michael et al (2000)). These sets of rules and systems defined for a translation in a language pair are hard-coded on the machine. The linguistic information used in an RBMT model is mainly the target and source languages collected from unilingual (one language), bilingual (two languages), or multilingual (more than two languages) dictionaries. In addition, the model also uses grammar covering the syntactic, semantic, and morphological regularities of each language. However, a well-built RBMT model requires highly skilled and expert human labour due to its complexity making it hard to build. In addition, the ambiguous properties of languages make them prone to take more time and efforts to resolve, especially in large and complex models. RBMT models require a lot of effort to be made functional in day-to-day life. Hence, the need for more efficient translation systems than RBMT still persists. EBMT methods make use of a large number of translation examples (John (2005)). Notably, EBMT models make use of bilingual corpora manipulation, i.e. the breaking down of a bilingual corpus into smaller parts, translating those parts into the target language, and recompiling it to form whole translated sentences. They do not account much for the syntax, semantic and morphological analysis of the target and source language (like RBMT models). In contrast, SMT is better when compared to RBMT and EBMT models, as it does not require human intervention (Adam (2008)). It is a way of translation wherein a statistical-based learning algorithm is applied to a large bilingual corpus that helps the machine learn the translation. This method also enables the machine to translate sentences not encountered by the machine during its training and testing. The objective of SMT is to convert an input word sequence from the source language into the target language. It has dominated academic MT research and a portion of the commercial MT sector in less than two decades. On the other hand, neural machine translation (NMT) is performed using a neural network (NN) (Stasimioti (2020)). Unlike SMT, NMT does not have a distinct translation model, language model, or model for reordering. Instead, it has a single sequence model that determines one word at a time. The prediction is based on the source sentence effort previously generated sequence in the target language. NMT is a deep learning-based method of machine learning that utilizes a large NN that relies on word vector representations.

Even though the NMT has achieved remarkable results in a few translation experiments using high-resource language, researchers are unsure if the NMT could actually replace SMT and if its success would extend to other tasks. Eventually, the experiment of (Michał (2016)) on the corpus of the United Nations (consisting of 15 low-resource languages) brings the fact. From the result of his experiment, it is evident that the performance of SMT is better than that of NMT for the majority of cases, as measured by BLEU score. Many researchers (Lohar et al (2019), Zhou et al (2017), Wang et al (2017), Castilho et al (2017)) have pointed out various disadvantages of NMT over SMT using low resource language, such as the fact that NMT requires more corpus and resources than SMT. In comparison with SMT, NMT training typically takes longer. Additionally, research has shown that when there is a domain incompatibility between testing and training data, SMT performance is superior to that of NMT (Xing et al (2018), Mahata et al (2018)). Long sentences are another area where SMT excels.

English and ILs are languages with less parallel text data, which motivates us to work with ILs. This research examines the effectiveness of SMT systems on low-resource language pairs, of which many are rarely worked on. The dataset used in our experiment for all fifteen Indian languages is tested for the first time for all languages using SMT. Hence, the objective of this workis to build an MT system using SMT for languages such as Assamese (AS),Malayalam (ML), Bengali (BN), Marathi (MR), Gujarati (GU), Kannada (KN), Hindi (HI), Oriya (OR), Punjabi (PA), Telugu (TE), Sindhi (SD), Sinhala (SI), Nepali (NE), Tamil (TA), and Urdu (UR) to English (EN) andvice versa and to check the effectiveness of SMT with low-resource language pairs.

Our main goal is to develop an MT system for low-resource languages, i.e., ILs, that can serve as a baseline system. The following is a summary of our work's main contributions:

- To the best of our knowledge, this work is the first attempt to use SMT with the Samanantar and OPUS Dataset to investigate the MT for all fifteen IL-EN and EN-IL pairs (both directions), including both the Dravidian and Indo-Aryan groups.

- To bring forth the linguistic approach of ILs in terms of translation. Scripts, writing style, and grammar with proper examples are also discussed.

- Various data filtration methods are investigated in order to clean the data and improve translation quality.

- Distance-based reordering is utilized to check the translation quality of ILs.

- Better realistic assessment of translation quality is possible from the presentation of results, as obtained using different automated metrics like BLEU, METEOR, and RIBES.

This paper is arranged as follows. Subsections 1.1 and 1.2 give some insight into SMT and cover the ILs used for our experiments. In Section 2, some prominent works on SMT and NMT using ILs are described. The experimental framework, including an overview of the dataset and methodology, is explained in Section 3. Section 4 narrates some of the prominent metrics used for MT evaluation. Results are presented in Section 5 followed by the conclusion and future direction in Section 6.

## 1.1 SMT

Statistical Machine Tramslation (SMT) is dependent on statistical methods (Philipp et al (2007), Richard et al (2002), Mary et al (2011) ). It is a data-driven technique that makes use of parallel-aligned corpora. It utilizes mathematical equations to calculate the likelihood of source-to-target language translation. Probability $P(Tl\ Si)$ is assigned by SMT. Here $Tl$ is the target language and $Si$ is the source input. It utilizes Bayes' theorem to determine the maximum probability $P(Tl|Si)$, which is as follows:

$$P(Tl \mid Si) \propto P(Tl)P(Si \mid Tl) \qquad (1)$$

SMT consists of three phases: the language model(LM) $P(Tl)$ for target language probability calculation, the translation model(TM) $P(Si|Tl)$ for conditional probability estimation of the target to the source language, and the decoder model (DM), which searches among possible source sentences the one which maximizes probabilities (Kumawat et al (2014)).
To calculate the probability of a sentence, the LM utilizes the $n$-gram model. It assigns the probability of a single word to the last $n$ words that come beforeit in the sentence and estimates the translation's likelihood. The chain rule aids in breaking down the sentence into conditional probability products.

$$\begin{aligned} P(s) &= P(w_1, w_2, w_3, \ldots, w_n) \\ &= P(w_1)P(w_2|w_1)P(w_3|w_1w_2)P(w_4|w_1w_2w_3)\ldots P(w_n|w_1w_2\ldots w_{n-1}) \\ &= P(w_1)P(w_2|w_1)P(w_3|w_1w_2)P(w_4|w_1w_2w_3)\ldots P(w_n|w_1w_2\ldots w_{n-k}) \end{aligned} \qquad (2)$$

Where, $P(s)$ is the probability of the sentence $s$, consisting of words $w_1$, $w_2$, ..., $w_n$, assuming a $k$-gram model. It utilizes the bilingual parallel corpus of the desired language pair. This is accomplished by calculating the likelihood of words or phrases extracted from sentences. The DM is the final and most crucial phase of SMT. It assists in the selection of words with the highest probability to be translated by maximizing the likelihood, i.e. $P(Tl)P(Si \mid Tl)$.

## 1.2 Language preference

India is a multilingual nation where people from various states use a variety of regional tongues. Such diversity of language brings difficulty in communicating with one another for information exchange. Further, limitations in public communication also bring inconvenience to share feelings, thoughts, opinions and facts, as well as to deal with business purposes. Moreover, there are many helpful resources available on the internet in English but many Indians struggle to take benefit of those due to language barriers. Hence, it is crucial to have an easy translation solution for regional languages to support effective communication and to help utilising global resources. To make it possible, technological innovation are continuing to find out efficient methods for a flawless translation using machines, because it is impractical to have human translators everywhere. For machine translation, an enormous amount of resources is required for training with a proper knowledge-base (rules) for better efficiency so as to fulfill the demand of a flawless translation solution. For translation, understanding the meaning of words is important, but words are not enough to constitute a language as a whole. They must be used in sentence construction that adheres to strict grammar rules and every language is having its own writing style. In our work, 15 commonly spoken languages (over various regions of India) are chosen. Table 1 describes the languages used in our experiments with their linguistic features (ethnologue (2022)). A short introduction about them in terms of translation is given below.

> *English(EN)*
> English language is the primary language of roughly 45 countries and is spoken by nearly 1,132 million people. It is written in Roman script, which uses both uppercase and lowercase characters. English uses the subject-verb-object structure. For example (expl1), "The poor man took food", and (expl2) "food took the poor man". When the position of the subject changes in the preceding sentences, the significance and meaning of the English sentence change.

### *Assamese(AS)*

Over 15 million native Assamese speakers live in the state of Assam in the northeastern region of India. It is one of Assam's official languages. Additionally, it is spoken in various regions of other northeastern Indian states. It uses the Bengali-Assamese script and is written left to right. It also follows the SOV format. "Gita is eating mango" is an English sentence that when translated into Assamese became গীতাই আম খাই আছ which follows subject object verb format. গীতাই (Gita, subject), আম(mango, object) and খাই আছ(is eating, verb).

### *Malayalam(ML)*

People in Kerala and a few societies in Karnataka and Tamil Nadu use Malayalam for communication. This language is spoken by about 35 million citizens. It uses the SOV style of writing and a nominative-accusative case marking sequence. It is written in Malayalam script in left-to-right fashion. Sentence like സീതയ്ക്ക് ചിത്രരചന ഇഷ്ടമാണ് which in English became "Sita loves drawing". Here the word സീതയ്ക്ക് (Sita,Subject), ഇഷ്ടമാണ് (loves,Verb) and ചിത്രരചന(drawing,Object).

### *Bengali(BN)*

It is the primary language of Bangladesh and the second most spoken language in India. Over 265 million people use it as their primary or second language. Approximately 11 million Bengali speakers exist in Bangladesh. In India, states such as Assam, Tripura, and West Bengal use this language. It is a member of the Indo-Aryan family. In Bengali sentences, the standard word order is Subject-Object-Verb. For example, in sentence রোজি ভাত খায় which in English is "Rosy eats rice ". Here রোজি (Rosy, Subject), ভাত (Rice, object) and খায় (Eats,Verb).

### *Marathi(MR)*

Marathi is associated with the Sanskrit-derived group of Indian languages and is used by 95 million people in India for communication, primarily in the central and western regions. The fourth most widely spoken language in India is Marathi, which has a sizable native-speaker population. Similar to Hindi and Nepali, Marathi is written in the Devanagari script in left-to-right order. It follows the Subject-Object-Verb order. For example, the sentence तो दूध िपतो, which means "He drinks milk." in English, has तो दूध िपतो where तो(He, subject), दूध (Milk, object), and िपतो(drinks, verb).

### Gujarati(GU)

Gujarati is spoken by 45 million citizens in Gujarat and is associated with the Indo-Aryan group. It uses the SOV writing style and is drafted from left to right in Gujarati script. For example, in the sentence તે આઇસ્ક્રીમ ખાય છે. which in English is "He is eating ice cream." where તે is subject, આઇસ્ક્રીમ is an object and verb is ખાય છે.

### Kannada (KN)

Karnataka's official language is Kannada, which is also widely used in other parts of India. In India, about 36 million people speak and write Kannada. Despite being a Dravidian language with extensive historical literature, Kannada has few computational linguistics resources, making it challenging to study the language's literature due to its semantic and syntactic diversity. Subject-Object-Verb is the way the Kannada language is structured. Kannada is a highly agglutinative language. It uses the left-to-right Kannada script For example, ರಾಮ ಶಾಲೆಗೆ ಹೋದ(SOV) is in English is "Rama went to school". Here, ರಾಮ(Rama, Subject), ಶಾಲೆ(school, object), and ಹೋದರು(went, verb).

### Hindi(HI)

Hindi is one of the official and national languages of India. There are more than 615 million people who use Hindi as their primary language, and even more than 341 million who speak it as a second language. However, the sentence structure is Subject Object Verb as shown in the example: गीता स्कूल जाती है। is in English is "Geeta goes to school".In this sentence, गीता (Gita ,Subject), स्कूल (School, Object) and जाती है(Goes, Verb). The Indian Constitution mandates that Hindi written in Devanagari be used as the Union's official language.

### Oriya(OR)

The Oriya language is the primary language of Odisha, a state in eastern India. Oriya belongs to the Eastern Indo-Aryan group of languages.

Its standard format is subject-object-verb (SOV) and is written in the Odia script from left to right.

### *Punjabi (PA)*

Punjabi text is written in a subject-object-verb format and is spoken in India and Pakistan, and a few small groups in the United Kingdom, United Arab Emirates, Malaysia, the United States, South Africa, and Canada. It is written in two scripts: the western Perso-Arabic Shahmukhi script and the eastern Gurmukhi script. Gurmukhi is drafted from left to right, whereas Shahmukhi is written in the opposite direction. ਅਸੀਂ ਭਾਰਤੀ ਹਾਂ is in English "We are Indians" where ਅਸੀਂ(We,Subject), ਭਾਰਤੀ (Are,Verb) and ਹਾਂ (Indians,Object).

### *Telugu(TE)*

Telugu is the official language of two Indian states in the south: Andhra Pradesh and Telangana. It is also spoken by the Telugu-speaking immigrant communities in the United States, Canada, and the United Kingdom. Text structure in Telugu takes the form of a subject-object-verb and from left to right.ఆమె నన్ను కొట్టింది in English "she beat me" where ఆమె(she, Subject), నన్ను(Me, Object) and కొట్టింది(beat, Verb).

### *Sindhi(SD)*

Sindhi is a language spoken by 25 million speakers in Pakistan and 5 million in India. It is written in a modified Perso-Arabic script in Pakistan (right-to-left), whereas it is written in a variety of scripts in India, like Devanagari, Khudabadi, and Gurmukhi (left-to-right). It follows the Subject-Object-Verb format. For example, the sentence "Partha loves books" is پارٿا ڪتابن سان پيار آهي where پارٿا(Partha, subject), ڪتابن سان (books, object) and پيار آهي (loves, verb).

### *Nepali(NE)*

It is official language and the lingua franca in Nepal, and also spoken by some communities in India. Nepali is written in left-to-right Devanagari script. It is a language written in Subject-Object-Verb order For example, "Sita ate apples" when converted to the Nepali language becomes सीताले स्याउ खाइन्. Here, सीताले(Sita, subject), स्याउ(apples, object) and खाइन्(ate, verb).

### *Sinhala(SI)*

The majority of Sri Lankans speak Sinhala as their first language. Sinhala is an Indo-Aryan language that differs from English in terms of grammatical structure, morphological variation, and subject-object-verb (SOV) word order. It is written in right-to-left Sinhala script. A sentence like "Pavan writes a letter" is in Sinhala is පවන් ලිපියක් ලියයි where පවන්(Pavan, subject), ලිපියක්(a letter, object) and ලියයි(writes, verb).

### *Tamil(TA)*

Tamil is a language spoken primarily in Tamil Nadu, a state in southern India, as well as in countries with a large Tamil diaspora, which includes Sri Lanka, Malaysia, and Singapore, to name a few. The phonological differences exist within Tamil Nadu between southern, western, and northern speech. Tamil is a Dravidian language of the southern branch, with a rich literary tradition dating back over 2000 years. Tamil spoken in India and Sri Lanka are two different dialects. It uses the Subject Object Verb format. For example sentence: "I like paintings" in Tamil becomes எனக்கு ஓவியங்கள் பிடிக்கும் where the Iஎனக்கு (I, Subject), விலங்குகள் (Paintings, Object) and பிடிக்கும் (Like, Verb).

### *Urdu(UR)*

It is Pakistan's national language and is also spoken widely in India. In Pakistan and India, Urdu is spoken by over 170 million citizens and is also spoken in some communities in the United Kingdom, the United States, and the United Arab Emirates. Script for Urdu is a modified and revised version of the Perso-Arabic script. Urdu writing structure is Subject Object Verb. For example "she reads a book" which in Urdu is وہ کتاب پڑھتی ہے where وہ(she, Subject), کتاب(book, object) and پڑھتی ہے(reads, verb).

Table 1: Linguistic Features of Languages Used in MT Experiments

| Languages | Script | Word Order | Family | Number of Speakers (in millions) | Writing Direction |
|---|---|---|---|---|---|
| Assamese (AS) | Bengali | SOV | Indo-European | 15 | left to right |
| Malayalam (ML) | Malayalam | SOV | Dravidian | 38 | left to right |
| Bengali (BN) | Bengali | SOV | Indo-European | 265 | left to right |
| Marathi (MR) | Devanagari | SOV | Indo-European | 95 | left to right |
| Gujarati (GU) | Gujarati | SOV | Indo-European | 60 | left to right |
| Kannada (KN) | Kannada | SOV | Dravidian | 36 | left to right |
| Hindi (HI) | Devanagari | SOV | Indo-European | 615 | left to right |
| Oriya (OR) | Oriya | SOV | Indo-European | 38 | left to right |
| Punjabi (PA) | Perso-Arabic, Gurmukhi | SOV | Indo-European | 125 | right to left left to right |
| Telugu (TE) | Telugu | SOV | Dravidian | 93 | left-to-right |
| Sindhi (SD) | Devanagari Perso -Arabic | SOV | Indo-European | 25 | left to right right to left |
| Sinhala (SI) | Sinhala | SOV | Indo-European | 17 | left to right |
| Nepali (NE) | Devanagari | SOV | Indo-European | 24 | left to right |
| Tamil (TA) | Tamil | SOV | Dravidian | 81 | left to right |
| Urdu (UR) | Urdu | SOV | Indo-European | 170 | right to left |
| English (EN) | Roman | SVO | Indo-European | 1,132 | left to right |

## 2 Related Work

A few works on SMT using some Indic Languages are discussed in this section.
(Dasgupta et al (2004)) has discussed a technique for English (EN) to Bengali (BN) MT that utilizes the syntax of EN sentences to BN while minimizing the time of translation. In the process to create the target sentences, a dictionary is used to know the object and subject, as well as other entities like person and number in their work.
English-to-Hindi (EN-HI) SMT system has been created by (Ananthakrishnan et al (2009)) using morphological and syntactic pre-processing in SMT model. In their work, the suffixes in HI language are segmented for morphological processing before rearranging the EN source sentences as per HI syntax.

In 2010, research has been conducted by (Zbib et al (2010)) at MIT, USA, using the grammatical structures in statistical machine translation with the Newswire corpus for Arabic to EN language to give better translation results. Work on Kannada-to-English MTS with SMT, by (Kumar et al (2015)), using Bible corpus on 20,000 sentences shows a remarkable feat with 14.5 BLEU score which is even supported by (Papineni et al (2002)).(Kaur et al (2011)) has presented a translation model based on SMT for English (EN) to Punjabi (PA) with their own corpus containing 3844 names in both languages with BLEU and word accuracy as 0.4123 (with range 0-1) and 50.22%, respectively.

(Nalluri et al (2011)) has created "enTel," an SMT-based EN to Telugu(TE) MT system, using the Johns Hopkins University Open Source Architecture (Li et al (2009)). For the purpose of training the translation system, TE par- allel dataset from the Enabling Minority Language Engineering (EMILLE)is used for their work.

In the year 2014, an SMT Framework for Sinhala(SI)-Tamil(TA) MT System has been created by (Randil et al (2014)). In their work, the result of SMT-dependent translation between language pairs, including TA-SI and SI-TA has been shown. Outcomes of the experiments using the SMT model give more noticeable results for the SI-TA than the TA-SI language pair. For languages closely related, SMT shows remarkable results.

In 2017, a survey has been conducted by (Khan et al (2017)) on the IL-EN language MT models reveal the importance of SMT over 8 languages i.e. Hindi (HI), Bengali (BN), Gujarati (GU), Urdu (UR), Telugu (TE), Punjabi (PA), Tamil (TA), and Malayalam (ML). In their work, EMILLE corpus (Nalluri et al (2011)) is used and Moses SMT model is preferred to make the translation models, with out-of-vocabulary (OOV) words transliterated to EN. In their work, the evaluation using BLEU, NIST and UNK counts as metrics reveals the overall SMT performance as satisfactory (PA-EN and UR-EN models as the best and the HI-EN and GU-EN models as the worst). An EN-BN SMT system has been presented by (Islam et al (2010)). In their work, to handle OOV (out-of-vocabulary) words, a transliteration module is presented. In order to address the systematic grammatical distinctions between EN and BN, a preposition handling module has been added. BLEU, NIST and TER scores has been used to check the effectiveness of their sys-tem.

Nowadays, NMT is widely appreciated for its advancement in the development of machine translation with remarkable improvement in quality. Hence, many researchers have compared both techniques for low and high-resource languages.

(Antonio et al (2017)) has performed a thorough evaluation using statistical-based and neural machine translation systems for nine language directions along a variety of dimensions. In their experiment, for long sentences, SMT systems perform better than the NMT. Recently, (Castilho et al (2017)) has used automatic metrics and expert translators to conduct a thorough quantitative and qualitative comparison of NMT and SMT. SMT shows better according to their experiments.

The comparison of NMT and SMT for the Nepali (NE) using the Nepali National Corpus (NNC) with 6535 sentences has been shown by (Acharya et al (2018)). The researchers have proved in their experiments that the SMT model performs better than the NMT-based system with a small corpus with a 5.27 BLEU score.

In 2021, Long Short-Term Memory networks (LSTMs) integrated with attention mechanism using WAT corpus have been used in experiments by (Singh et al (2003)) to achieve a 15.7 BLEU score as opposed to a baseline of 14.5 BLEU score.

(Abujar et al (2021)) has developed a BN-EN MT model on AmaderCAT corpus using Sequence-to-Sequence (seq2seq) architecture, a special class of Recurrent Neural Networks to develop the translation system and has achieved a BLEU score of 22.3.

In the year 2021, translation of English and Hindi-to-Tamil languages us- ing both SMT and NMT has been presented by (Akshai et al (2021)). The disadvantages of NMT have been shown in their experiments such as the occurrence of numerous errors by NMT when interpreting domain terms and OOV (Out of vocabulary) phrases. NMT frequently constructs inaccurate lexical choices for polysemous words and occasionally counters reordering mistakes while translating words and domain terms. The translations that have been generated by the NMT models mostly include repetitions of previously transcribed words, odd translations, and many unexpected sentences having no correlation with the original sentence.

# 3 Experimental Framework

## 3.1 Dataset

Samanantar and OPUS datasets for model building and standard benchmark dataset i.e. Flores 200 for testing are utilized. Samanantar is the largest corpus collection for ILs (Gowtham et al (2022)). The collection includes more than 45 million sentence pairs in English and 11 ILs. The Samanantar Corpus has been used for Assamese (AS), Malayalam (ML), Bengali (BN), Marathi (MR), Gujarati (GU), Kannada (KN), Hindi (HI), Oriya (OR), Punjabi (PA), Telugu (TE), and Tamil (TA) for the experiments. OPUS is a large resource with freely available parallel corpora. The corpus includes data from many domains and covers over 90 languages (Tiedemann (2012)). The OPUS corpus is used for Sinhala (SI), Sindhi (SD), Urdu (UR), and Nepali (NE). Table 2 gives statistics of the dataset used in our experiments.
FLORES-200 (Marta et al (2022)) dataset is a multilingual parallel dataset with 200 languages, that are used as human-translated benchmarks. It consists of two corpora, labeled "dev" (997 lines) and "devtest" (1013 lines). The "dev" dataset has been used for fine-tuning, and the "devtest" dataset has been used for testing.

## 3.2 Methodology

Our proposed process comprises of following major steps:

1. **Setting up SMT System** Moses SMT Toolkit is used to build our SMT systems. It is written in C++ and Perl. At the moment, thisis one of the best SMT tools available. First, Moses, GIZA++ (Och (2003)), CMPH (for binarization) and SRILM in Ubuntu are installed. For training, fine-tuning and testing processes, the system needs a parallel corpus of the language pair in addition to configurable phases according to developer's choice to follow.

2. **Data Preprocessing** A qualitative corpus plays a major role in any MT task. While obtaining corpora from various sources, data qual- ity i.e. critical for the effectiveness of an MT system, can never be ascertained. So, removing unnecessary noise is an important task before using the data to train our statistical machine translation model. Following processes are used to preprocess and clean it:

Table 2: Parallel corpus statistics

| English to Indic | Parallel Corpus(Sentences) |
|---|---|
| Assamese (AS) | 0.14M |
| Malayalam (ML) | 5.85M |
| Bengali(BN) | 8.52M |
| Marathi(MR) | 3.32M |
| Gujarati(GU) | 3.05M |
| Kannada(KN) | 4.07M |
| Hindi(HI) | 8.56M |
| Oriya(OR) | 1.00M |
| Punjabi(PA) | 2.42M |
| Telugu(TE) | 4.82M |
| Sindhi(SD) | 1.95M |
| Sinhala(SI) | 8.68M |
| Nepali(NE) | 3.35M |
| Tamil(TA) | 5.16M |
| Urdu(UR) | 8.95M |

- **Data Cleaning and Formatting** The goal of data cleaning is either to find and fix or to delete erroneous data from the corpus. Here, characters those are used neither in ILs nor in English are removed. Some of the punctuation in extended Unicode is converted to its standard counterpart. Numbers in the IL corpus are converted from English to IL scripts. Characters outside the standard alphabets of the language pair, extra spaces, and unprintable characters are also removed from the corpus. The preprocessing techniques used in our work have been summarized as follows:
    - Removing unprintable characters
    - Removing characters outside the language pair
    - Removing extra spaces
    - Deaccenting accented characters
    - Changing non-standard Unicode punctuation characters in both corpora to their standard counterparts
    - Changing uncommon punctuations to more common ones
    - Changing numbers to a uniform numbering system and script

3. **Tokenization:** It is the process of dividing a character sequence into smaller units known as tokens based on a given character sequence and a specified document unit. Words, punctuation, and numerals serve as these tokens in our instance. The corpus is tokenized using a modified Moses tokenizer (Koehn et al (2007)). Redundant punctuations (quotation marks, apostrophes, and commas) are also removed from the corpus.

4. **Training Truecasing Model:** This is the procedure for adding case information to text that has been incorrectly cased or is not cased (Lita et al (2003)). Data sparsity is lessened with the use of true casing. A truecaser model (a model which changes the words at the beginning of the sentence to the most common casing) is trained on the training dataset. The Moses truecasing is used for the same.

5. **Training Language and Translation Models**: In MOSES, the training procedure utilizes word and segment occurrences to draw connections between the target and source languages. The language and translation models are trained on the training dataset and binarized. GIZA++ grow-diag-final-and alignment is used for word alignments, which start with the intersection of the two alignments and then add the additional alignment points.
   The grow-diag-final-and model starts with the intersection of the alignments from source to target and target to source, then two steps are used to add additional alignment points (Och (2003)):

   *grow-diag*: For every neighboring point to the alignments measured, if either source or target word is not aligned already but is present in the union of the alignment, then the neighboring point is included in the alignment.

   *final*: If any phrase pairs are unaligned but present in the union, add the point to the alignment.

   - **Word Alignment Model:** After preprocessing the words, the next step is word alignment. The proposed work employs the GIZA++ (Och (2003)) incorporation of the IBM models to accomplish the word procedure. The GIZA++ model assesses the likelihood of word-to-word alignment for each source and target word in each sentence. To produce a good-quality word alignment, the alignment is produced using a series of successive estimations. To process a corpus with a larger quantity of sentences, the process takes several hours. The alignment method's outcomes establish a connection between the target and source words.

- **Reordering** It is the process of restructuring the word order of one natural language sentence to make it more similar to the word order of another natural language sentence. It is a critical task in transcription for languages with different syntactic structures. The Moses system learns different reordering possibilities for each phrase during the training process. Instead of default reordering, the model uses the distance reordering model (Kumawat et al (2014)).

    - **Distance-Based Reordering:** The reordering of the tar- get output phrases is represented by the relative distortion probability distribution $re$ ($S_t$, $E_{t-1}$). Here, $S_t$ refers to the starting position of the source phrase that is interpreted into the $t-1$ th target phrase. The reordering distance ($S_t$ -$E_{t-1}$) is calculated as follows: When taking source words out of sequence, the reordering distance is the number of words ignored (either forward or backward). If two phrases are translated in sequence, then $t = E_{t-1}+1$; that is, the first word of the phrase immediately follows the last word of the previous phrase. A reordering cost of $re(0)$ is used in this case. The distance-based model assigns a linear cost to reordering distance, implying that the movement of phrases over long distances is more expensive.

6. **Fine tuning**: It is the process of determining the best configuration file settings for a translation model when it is used for a specific pur- pose. It uses a translation model to translate all 15 ILs source language phrases in the tuning set. Then, it compares the model's output to a set of reference (human translations) and adjusts the settings to improve translation quality. This procedure is repeated several times. The tuning process repeats the steps with each iteration until the translation quality is optimized. The model is fine-tuned on the preprocessed Flores-200 dev dataset.

7. **Translation**: The final model is used to translate the preprocessed

Flores-200 devtest dataset from the source to the target language.

8. **Postprocessing and Detokenization**: Redundant punctuation marks (quotation marks, apostrophes, and commas) are removed, and the translation file is detokenized using the Moses detokenizer.

9. **Evaluation**: The evaluation metrics use for our experiments are METEOR (Banerjee et al (2005)), RIBES (Wołk et al (2016)), and BLEU (Papineni et al (2002)).

# 4 Essential metrics for MT translation evaluation

The most crucial phase of any MT system is MT evaluation. Both automatic and manual methods can be applied to analyze MT tasks. The effectiveness of a system's output can be evaluated either directly through human assessments, or indirectly using reading cases, other downstream activities, and even through estimating the amount of effort necessary to rectify the output. A better outcome is obtained through manual evaluation, which includes task-based evaluations, fluency and adequacy scores, human vot- ing for translations task, post-editing measures, etc. However, the major challenges of manual evaluation are time-intensiveness, absence of repeatability and high cost. In order to evaluate the effectiveness of MT output, different automated approaches are there such as Metric for Evaluation of Translation with Explicit Ordering (METEOR), Bilingual Evaluation Understudy(BLEU), Levenshtein, Rank-based Intuitive Bilingual Evaluation Score(RIBES), Word Error Rate (WER) and NIST exist. Several intuitive advantages exist for automated metrics that can give points for synonyms or paraphrases. A few of the evaluation metrics which are used in our work are discussed below

1. **Bilingual Evaluation Understudy (BLEU):** The most widely used method for evaluating machine translation (MT) is known as BLEU. This method, first introduced in 2002 (Papineni et al (2002)) exam- ines one or more reference translations to the hypothetical translation. When the hypothetical translation matches numerous strings with the reference translation, the MT evaluation gives it a higher score. The BLEU system assigns a translation a score from 0 to 1. However, it is usually represented as a percentage value. The nearer the translation is to 1, the more it corresponds to the reference translation. This matching of translation is conducted word-by-word in the same word order in both datasets. SacreBLEU is used to calculate the BLEU scores of baseline models.

2. **Rank-based Intuitive Bilingual Evaluation Score (RIBES):** It is calculated by incorporating a rank correlation coefficient before unigram matches, eliminating the necessity for higher-order n-gram matches. This metric is concerned with word order. To compare SMT and reference translations, it employs Kendall's tau coefficient ($\tau$) based on word order to indicate rank differences (Wołk et al (2016)). To assure positive values, the coefficient is normalized as shown below:

$$\text{Normalized Kendall's } \tau \text{ (NKT)} = \frac{\tau + 1}{2} \quad (5)$$

This coefficient can be paired with unigram-precision $p_1$ and Brevity Penalty $BP$ and changed to prevent overestimation of the correlation between only relevant words in SMT and reference translations.

$$\text{RIBES} = \text{NKT}.(p_1^\alpha).(BP^\beta) \quad (6)$$

Here, $\alpha$ and $\beta$ are parameters between 0 and 1.

3. **Metric for Evaluation of Translation with Explicit Ordering (METEOR):** Meteor scores a translation depending on explicit word-by-word similarities between both the translation and a provided reference translation (Banerjee et al (2005)). It is specifically created to generate sentence-level scores that are highly correlated with human evaluations of translation quality. Meteor utilizes and highlights recall in combination with precision, a feature that numerous measures have verified as crucial for a strong correlation with human judgments. It also intends to address the problem of imprecise reference translations by utilizing adaptable word matching in consideration with synonyms and morphological variances. To achieve a score of 1, the words of the machine-generated output should be present in the reference and each of the words of the reference is in the machine-generated output.

# 5 Results and Discussion

In this work, the evaluation metrics used are METEOR (Banerjee et al (2005)), RIBES (Wołk et al (2016)), and BLEU (Papineni et al (2002)). All the evaluation metrics used in our work are prominent metrics for determining the quality of the machine-translated text.

Table 3 displays the translation of all the 15 ILs to English and vice versa using SMT without fine-tuning. Evaluation metrics of SMT with finetuning using the Flores-200 dev dataset are shown in Table 4. RIBES and METEOR range is 0-1. For EN-IL and IL-EN language using SMT, the BLEU score lies between 0.46 to 13.09 and 0.49 to 15.41 respectively. The RIBES score for EN-IL and IL-EN is between 0.04 to 0.63 and 0.14 to 0.61 respectively. METEOR scores lie between 0.01 to 0.28 for EN-IL and 0.02 to 0.28 for IL-EN. SMT models using distance reordering techniques are giving better BLEU Scores for languages BN, PA, UR, HI, and GU than the rest. Without fine-tuning, SI performs the worst in terms of all three metrics of all languages in both directions, whereas with fine-tuning EN-SI and TA-EN perform worse than all other EN-IL and IL-EN models respectively with all three metrics. HI and BN languages have qualitative, large, and less noisy datasets compared to other languages. Hence, HI performs the best among all languages without fine-tuning in all three metrics in both directions, and BN performs the best among all languages with fine-tuning in both direc- tions with respect to BLEU and RIBES. In addition, UR and PA also produce good RIBES metrics than other languages. RIBES score for PA is 0.63(forEN-PA) and 0.61(PA-EN), and for UR, RIBES score is 0.62(EN-UR) and 0.61(UR-EN).

Table 3: Evaluation Metrics Result of SMT without Finetuning

| Languages | Pairs | BLEU | RIBES | METEOR |
|---|---|---|---|---|
| **AS** | EN-AS | 1.90 | 0.50 | 0.09 |
|  | AS-EN | 3.21 | 0.46 | 0.11 |
| **ML** | EN-ML | 3.79 | 0.27 | 0.08 |
|  | ML-EN | 4.59 | 0.43 | 0.12 |
| **BN** | EN-BN | 6.41 | 0.62 | 0.17 |
|  | BN-EN | 3.06 | 0.45 | 012 |
| **MR** | EN-MR | 3.17 | 0.43 | 0.09 |
|  | MR-EN | 3.62 | 0.43 | 0.09 |
| **GU** | EN-GU | 7.62 | 0.56 | 0.16 |
|  | GU-EN | 10.14 | 0.59 | 0.21 |
| **KN** | EN-KN | 5.06 | 0.40 | 0.11 |
|  | KN-EN | 7.17 | 0.51 | 0.16 |
| **HI** | EN-HI | 13.09 | 0.63 | 0.28 |
|  | HI-EN | 15.41 | 0.64 | 0.28 |
| **OR** | EN-OR | 3.92 | 0.59 | 0.14 |
|  | OR-EN | 6.41 | 0.52 | 0.17 |
| **PA** | EN-PA | 7.22 | 0.63 | 0.18 |
|  | PA-EN | 11.7 | 0.61 | 0.24 |
| **TE** | EN-TE | 8.16 | 0.42 | 0.12 |
|  | TE-EN | 5.77 | 0.52 | 0.18 |
| **SD** | EN-SD | 1.29 | 0.39 | 0.08 |
|  | SD-EN | 2.48 | 0.35 | 0.09 |
| **SI** | EN-SI | 0.93 | 0.05 | 0.02 |
|  | SI-EN | 0.49 | 0.14 | 0.05 |
| **NE** | EN-NE | 6.00 | 0.58 | 0.16 |
|  | NE-EN | 8.29 | 0.53 | 0.19 |
| **TA** | EN-TA | 2.78 | 0.16 | 0.05 |
|  | TA-EN | 2.64 | 0.31 | 0.07 |
| **UR** | EN-UR | 9.43 | 0.62 | 0.24 |
|  | UR-EN | 11.35 | 0.61 | 0.23 |

Table 4: Evaluation Metrics Result of SMT with Finetuning

| Languages | Pairs | BLEU | RIBES | METEOR |
|---|---|---|---|---|
| **AS** | EN-AS | 2.17 | 0.50 | 0.08 |
| | AS-EN | 3.21 | 0.42 | 0.10 |
| **ML** | EN-ML | 2.05 | 0.23 | 0.06 |
| | ML-EN | 1.84 | 0.27 | 0.06 |
| **BN** | EN-BN | 8.26 | 0.63 | 0.19 |
| | BN-EN | 12.16 | 0.60 | 0.23 |
| **MR** | EN-MR | 2.43 | 0.39 | 0.08 |
| | MR-EN | 2.49 | 0.36 | 0.07 |
| **GU** | EN-GU | 5.82 | 0.52 | 0.14 |
| | GU-EN | 3.56 | 0.45 | 0.01 |
| **KN** | EN-KN | 3.35 | 0.09 | 0.14 |
| | KN-EN | 3.67 | 0.41 | 0.10 |
| **HI** | EN-HI | 8.64 | 0.57 | 0.22 |
| | HI-EN | 5.38 | 0.49 | 0.14 |
| **OR** | EN-OR | 5.25 | 0.58 | 0.15 |
| | OR-EN | 2.22 | 0.39 | 0.11 |
| **PA** | EN-PA | 5.71 | 0.60 | 0.15 |
| | PA-EN | 7.75 | 0.55 | 0.19 |
| **TE** | EN-TE | 4.4 | 0.38 | 0.10 |
| | TE-EN | 3.34 | 0.44 | 0.12 |
| **SD** | EN-SD | 1.59 | 0.41 | 0.09 |
| | SD-EN | 2.53 | 0.38 | 0.09 |
| **SI** | EN-SI | 0.46 | 0.04 | 0.01 |
| | SI-EN | 3.11 | 0.37 | 0.11 |
| **NE** | EN-NE | 4.00 | 0.55 | 0.14 |
| | NE-EN | 5.25 | 0.49 | 0.13 |
| **TA** | EN-TA | 1.86 | 0.16 | 0.05 |
| | TA-EN | 1.03 | 0.08 | 0.02 |
| **UR** | EN-UR | 6.34 | 0.56 | 0.19 |
| | UR-EN | 7.07 | 0.54 | 0.18 |

Even though SI has a good amount of corpus, the corpus does not have reli- able translations compared to other languages. For example, the sentence in English "Heb. 11:32-34; Judg. 16:18-21, 28-30 Jehovah's spirit operated on Samson in a unique way because of unusual circumstances" has been trans- lated to Sinhala in the corpus as "11:32-34; ගති.", which only translates "Heb. 11:32-34;". Hence, SI does not perform well compared to other lan- guages. Similarly, in the EN-TA corpus, the sentence "He's my boss" has been translated to "அவர் எனது மேலாளர் மட்டும்தான்." which ac- tually means "He is only my manager". From the example, it is clear that EN-TA corpus also has ambiguity. Additionally, even though the ILs-English and English-ILs systems are trained using the same corpus, a significant dis- crepancy in the BLEU scores is observed. This is due to the significant morphological diversity of ILs and the relative difficulty of translating from English to ILs. It has been observed that SI has a high number of lines (8.68 M) but performs poorly as compared to languages like PA (2.42 M) and GU (3.05 M). It is also observed that languages with very steep slopes tend to have low scores. For example, EN-TA and EN-ML have 60% sentences with less than 4 tokens, and they have not-so-good scores as shown in Figure 1. In contrast, languages with good scores, like HI and BN have more gentle slopes. So, length of sentences is a contributing factor. EN-SD is an ex- ception which has a gentle slope but does not give good scores, because the corpus does not have good translation quality. Therefore, the quality of the corpus matters more than the size of the dataset.

# 6 Conclusion and Future Work

This paper has presented the MT work for 15 ILs to English and vice versa using SMT. It also describes the linguistic features of all 15 ILs. A tailor-made preprocessing approach has been incorporated into this work. The model has utilized the grow-diag-final-and alignment model and distance reordering model. For checking the quality of translation, different evaluation Metrics such as BLEU, RIBES, and METEOR are utilized in this work. From the result, it is observed that the proposed SMT model is quite satisfactory for some of the ILs. However, the level of performance is not at par with the rest of the ILs and there lies the need of improvement to be made. Due to the scarcity and quality of parallel corpus, the metrics obtained are quite low.

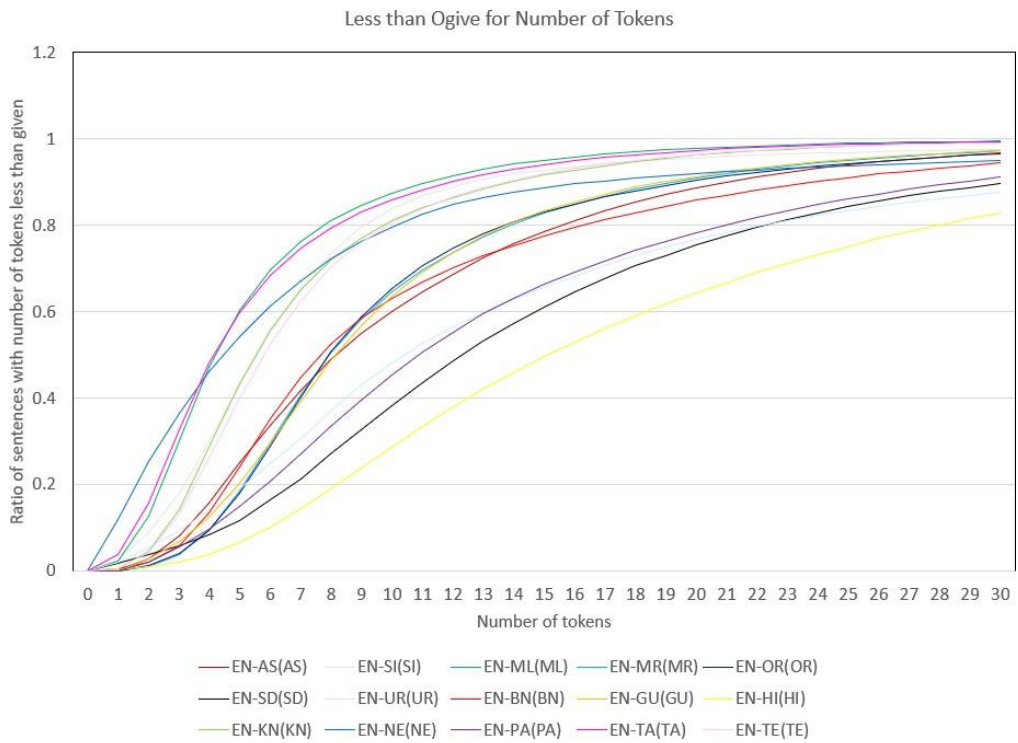

Figure 1: Less than ogive for number of tokens in a sentence for all fifteen language corpora

It has been observed that the translations of some of the languages are not sufficiently accurate. Measures of validating corpus quality shall be explored in order to observe the corpus quality and remove inaccurate lines. Dravid- ian languages are in general agglutinative languages (words are made up of morphemes, with each morpheme contributing to the meaning of the word). In future, means to infer translations from the breakdown of words in these languages shall also be explored.

Interestingly, in some of the ILs, finetuning schemes are hampering the quality. The causes of this phenomenon shall be analyzed and mitigated via techniques such as noise reduction, corpus cleaning, and finetuning schemes for those languages to ensure better quality. In addition, more language pairs and corpora can be analyzed and evaluated using various other methods. Other techniques, such as hybridized SMT-NMT systems and the usage of other alignment and reordering models will be studied for further course of research.